\documentclass[runningheads]{llncs}
\usepackage[T1]{fontenc}
\usepackage{cite}
\usepackage{amsmath,amssymb,amsfonts}
\usepackage{algorithm}
\usepackage{algpseudocode}
\usepackage{graphicx}
\usepackage{caption}
\usepackage{textcomp}
\usepackage{xcolor}
\usepackage{hyperref}
\usepackage{color}

\usepackage{comment}
\usepackage{balance}

\begin{document}
\title{Beyond ZOH: Advanced Discretization Strategies for Vision Mamba}
%
\author{Fady Ibrahim\orcidID{0009-0004-6534-598X} \and Guangjun Liu\orcidID{0000-0002-3301-1166} \and Guanghui Wang\orcidID{0000-0003-3182-104X}}
\authorrunning{F. Ibrahim et al.}
%
\institute{Toronto Metropolitan University, 350 Victoria St, Toronto, ON M5B 2K3}
\maketitle 
\begin{abstract}

Vision Mamba, as a state space model (SSM), employs a zero-order hold (ZOH) discretization, which assumes that input signals remain constant between sampling instants. This assumption degrades temporal fidelity in dynamic visual environments and constrains the attainable accuracy of modern SSM-based vision models. In this paper, we present a systematic and controlled comparison of six discretization schemes instantiated within the Vision Mamba framework: ZOH, first-order hold (FOH), bilinear/Tustin transform (BIL), polynomial interpolation (POL), higher-order hold (HOH), and the fourth-order Runge–Kutta method (RK4). We evaluate each method on standard visual benchmarks to quantify its influence in image classification, semantic segmentation, and object detection. Our results demonstrate that POL and HOH yield the largest gains in accuracy at the cost of higher training-time computation. In contrast, the BIL provides consistent improvements over ZOH with modest additional overhead, offering the most favorable trade-off between precision and efficiency. These findings elucidate the pivotal role of discretization in SSM-based vision architectures and furnish empirically grounded justification for adopting BIL as the default discretization baseline for state-of-the-art SSM models. 


\keywords{Mamba \and State Space Models \and Vision Mamba \and Discretization \and Bilinear Transform \and Higher-Order Hold.}
\end{abstract}

\section{Introduction}\label{sec:intro}

State Space Models (SSMs), such as Vision Mamba (ViM), have received significant attention in recent years \cite{gu2023mamba}\cite{liu2024vmambavisualstatespace}.
ViM presents a promising alternative to Vision Transformers (ViTs), which leverage their ability to capture complex dependencies, but at a quadratic rate of calculations due to their attention mechanisms~\cite{dosovitskiy2020vit}\cite{ibrahim2025survey}\cite{zhang2025mix}. SSMs have near-linear scalability in sequence length and hardware-efficient processing ~\cite{gu2023mamba}. Mamba has recently been adapted to computer vision tasks; ViM~\cite{zhu2024vision}, Vmamba \cite{liu2024vmambavisualstatespace}, and VideoMamba ~\cite{li2025videomamba} demonstrate Mamba's ability to handle long-range dependencies while addressing the computational and memory bottlenecks faced by Transformer-based models in the visual domain.
Despite their efficiency, Mamba models used for vision inherit a limitation from classical control theory, the use of zero-order hold (ZOH) for continuous-to-discrete conversion\cite{zhang2007comparisonzohfoh}. ZOH assumes that input values remain constant between samples—an assumption that oversimplifies visual data, where spatial signals exhibit continuous variation, sharp edges, textures, and structured frequency content. While ZOH simplifies implementation and guarantees stability, it introduces discretization lag and truncation errors that accumulate across long sequences of image patches or tokens.

To address these limitations, this work systematically investigates a controlled and hardware-aware study of advanced discretization strategies that better capture continuous dynamics within Mamba-based visual systems. We implement and evaluate six discretization methods within the ViM framework:

\begin{enumerate}
    \item Zero-Order Hold (ZOH) – The baseline method used in original ViM implementations, serving as a reference point for comparison \cite{zhang2007comparisonzohfoh}.
    \item First-Order Hold (FOH) – A linear interpolation method that models smoother transitions between samples \cite{zhang2007comparisonzohfoh}.
    \item Bilinear/Tustin Transform (BIL) – A signal processing technique that converts stable analog filters into equivalent digital filters, allowing analog designs to be implemented digitally, preserving stability \cite{aastrom2013compcontrolsystems}\cite{oppenheim2010discretetimesig}.
    \item Polynomial Interpolation (POL) – A higher-order approximation that fits polynomial curves through sample points, improving temporal smoothness \cite{chen2025approxmethodsdiscretize}\cite{ascher1998computermethodsode}.
    \item Higher-Order Hold (HOH) – An extension of FOH using higher-degree polynomials to better approximate complex temporal variations \cite{takacs2022hohmethods}.
    \item Runge–Kutta 4 (RK4) – A fourth-order integration method providing high-precision discretization for systems with fast-changing dynamics \cite{butcher1964rk4processes}.
\end{enumerate}

By integrating these methods within a unified CUDA-based execution framework that preserves identical memory access patterns, kernel fusion, and scan-based recurrence across methods, we analyze how each affects model accuracy, training stability, and generalization across benchmark datasets such as ImageNet-1k \cite{deng2009imagenetdb}, Cifar100 \cite{krizhevsky2009cifar}, ADE20K \cite{zhou2019semanticade20k}, and MS COCO \cite{lin2014microsoftcocodataset}. 

This work establishes a foundational understanding of how numerical methods, often overlooked in architectural design, directly influence the performance and stability of modern state space visual models. This work establishes discretization as a first-class design decision in ViM and provides the first hardware-consistent benchmark of numerical integration methods for state-space vision models. Although the work in this paper is based on ViM, the proposed discretization methods can be directly applied to other SSMs as well. 
The contributions of this paper are threefold: 

\begin{enumerate}
    \item Identifying the limitations introduced by ZOH in current ViM and formally characterizing its impact on visual modeling.
    \item Implementing and evaluating five advanced discretization strategies (six including ZOH) within the ViM framework across large-scale classification, segmentation, and detection benchmarks.
    \item Demonstrating that while POL and HOH methods yield the highest accuracy, BIL provides a balanced improvement in precision and efficiency.
\end{enumerate}



\section{Related Work}

In foundational SSM frameworks, the conversion from continuous to discrete time uses a fixed discretization rule. This ensures that the model can handle continuous input $x(t)$ by propagating information through hidden states $h(t)$.
Modern SSM architectures, such as Vision Mamba (ViM) \cite{zhu2024vision}, VMamba \cite{liu2024vmambavisualstatespace}, and its derivatives \cite{huang2024localmamba}\cite{yang2024plainmamba}\cite{xu2024vimsurvey}\cite{zhang2024vismambasurvey} operate using this structured fixed discretization step.
The fundamental SSM layer introduced in \cite{smith2022simplifiedssl} relies on this continuous-to-discrete system translation. The original Mamba research \cite{gu2023mamba}, while focusing on content-dependent selection, maintains that SSMs were originally defined as discretizations of continuous systems.

The standard and most commonly cited discretization technique across modern SSM architectures is the ZOH method. The ZOH method is predicated on the assumption that the input remains constant over a given time interval $\Delta$. This method transforms the continuous system parameters, such as the evolution matrix $\boldsymbol{A}$ and input matrix $\boldsymbol{B}$, into their discrete counterparts. ZOH discretization is utilized implicitly in many prominent SSM models and extensions, including Mamba \cite{gu2023mamba}, Structured State Space Sequence (S4) models \cite{zhang2024vismambasurvey}\cite{xu2024vimsurvey}, VMamba \cite{liu2024vmambavisualstatespace}, and VideoMamba \cite{li2025videomamba}.

The general framework of State Space Models allows for various discretization methods. One line of work explored simplifying the complexity of state matrices themselves, while later research focused on integrating an alternative discretization method beyond ZOH to enhance model expressivity \cite{lahoti2026mamba3}\cite{rahman1990trapezoidalrule}. This is similar in concept to BIL (Tustin's method) \cite{oppenheim2010discretetimesig}:
\begin{itemize}
    \item Real Valued Simplification\cite{smith2022simplifiedssl}: The Mega model, a derivative of S4, represented the discretization of continuous system parameters as an Exponential Moving Average (EMA) term, which led to a simplification of the SSM model that only focused on real-valued parameters, reducing computational cost.
    \item Trapezoidal Discretization (Mamba-3)\cite{lahoti2026mamba3}: The work experimented with the use of trapezoidal discretization (trapezoidal/Euler's rule) \cite{rahman1990trapezoidalrule} as an improvement to the core SSM components within Mamba for large language modeling. The complex valued SSMs, when discretized using the general trapezoidal rule, yielded an equivalent real valued SSM equipped with data-dependent Rotary Positional Embeddings (RoPE) \cite{heo2024rope-rotaryposencode}\cite{movahedi2025selectiverope2}. The successful application of this method resulted in precise algorithmic reasoning, with which previous SSM variants struggled. \cite{lahoti2026mamba3}.
\end{itemize}

In summary, prior work on state space models for vision has largely treated discretization as a fixed and secondary design choice, with ZOH implicitly adopted across most Mamba-based architectures. Recent efforts, such as real-valued simplifications and trapezoidal discretization, demonstrate that revisiting numerical formulations can improve specific capabilities; however, these approaches are either task-specific, limited to language modeling, or evaluate a single alternative in isolation. In contrast, this work systematically examines discretization itself as a first-class modeling component, providing a comparison of five advanced discretization methods within a consistent ViM framework. By isolating discretization from other architectural changes, our study clarifies its independent impact on performance in visual tasks, establishing discretization as a key design choice rather than a fixed implementation detail.

\section{Methodology}
\subsection{Preliminaries}\label{sec:prelim}

\subsubsection{Mamba Selective State Space Model (S6)} 

Mamba enhances the context-aware capabilities of traditional SSMs by making its parameters functions of the input. The original Mamba architecture was intended for 1D sequential data such as language tokens. SSMs map a 1D function $x(t) \in \mathbb{R}$ to an output $y(t) \in \mathbb{R}$ through a hidden state $h(t) \in \mathbb{R}^N$.
Using the parameters 
$\mathbf{A} \in \mathbb{R}^{N \times N}$$, \mathbf{B} \in \mathbb{R}^{N \times 1}$$, $$\mathbf{C} \in \mathbb{R}^{1 \times N}$, the continuous system is defined by

\begin{equation}
\begin{aligned}
\frac{dh(t)}{dt} = \mathbf{A}h(t) + \mathbf{B}x(t), \quad y(t) = \mathbf{C}h(t)
\end{aligned}
\label{eq:ssm}
\end{equation}

For the discrete sequences defined by the input \( x_t = (x_0, x_1, \dots) \in \mathbb{R}^L \), the continuous system parameters in Eq.~\eqref{eq:ssm} must be discretized using a step size \( \Delta \). ZOH~\cite{zhang2007comparisonzohfoh} is used as the simplest discretization method in~\cite{zhu2024vision, li2025videomamba, liu2024vmambavisualstatespace}. Using step size \( \Delta \), the continuous system parameters \( \mathbf{A}, \mathbf{B} \) are converted to \( \bar{\mathbf{A}}, \bar{\mathbf{B}} \) as

\begin{equation}
\begin{aligned}
    \bar{\mathbf{A}} &= e^{\mathbf{A}\Delta}, \\
    \bar{\mathbf{B}} &= (\mathbf{A}\Delta )^{-1} (e^{\mathbf{A}\Delta} - \mathbf{I}) \cdot \Delta \mathbf{B}
\end{aligned}
\label{eq:zoh}
\end{equation}
and the discretized system equation can be represented as
\begin{equation}
\begin{aligned}
h_t = \bar{\mathbf{A}}h_{t-1} + \bar{\mathbf{B}}x_t, \quad y_t = \mathbf{C}h_t
\end{aligned}
\label{eq:ssmdisc}
\end{equation}

The innovation of Mamba lies in the Selective Scan Mechanism (S6) ~\cite{gu2023mamba}, making these SSM parameters $\mathbf{B}$, $\mathbf{C}$, $\Delta$ functions of the input.
The original Mamba block ~\cite{gu2023mamba} introduced a hardware-aware algorithm to efficiently compute the discretization and selective SSM parameters in Eq.~\eqref{eq:ssm} and Eq.~\eqref{eq:ssmdisc} to compute the latent hidden state and output. This mechanism allows Mamba to selectively retain or discard information based on the relevance of the input sequence.

\subsubsection{ViM Model}
\begin{figure}[t]
\centering
\includegraphics[width=0.85\linewidth]{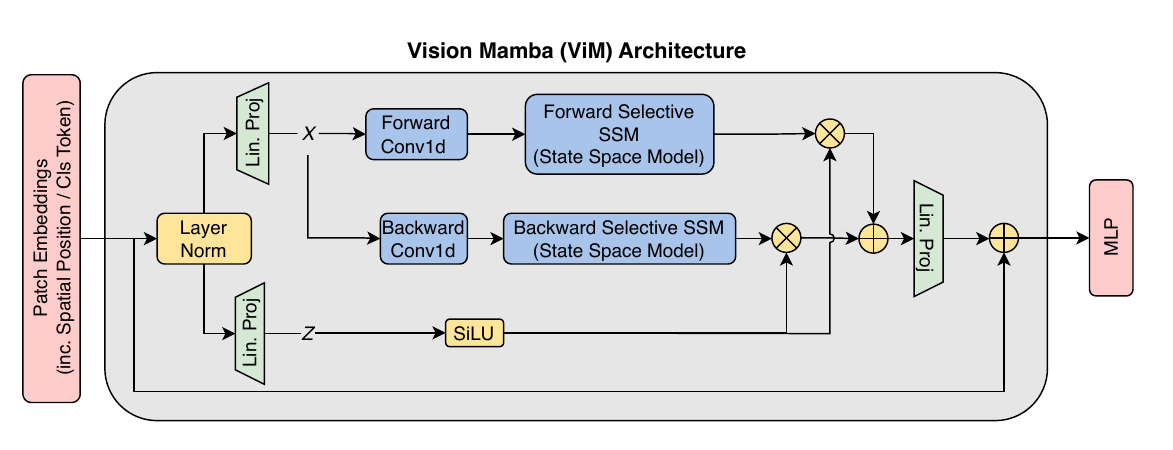}
\caption{The architecture of Vision Mamba model.}
\label{fig1}
\end{figure}

ViM extends the Mamba architecture, as shown in Fig.\ref{fig1}, to address the complexities of visual data. It addresses position sensitivity by incorporating position embeddings, similar to those used in ViT ~\cite{dosovitskiy2020vit}. The original Mamba block's unidirectional scanning suits causal sequences. However, visual data are non-causal. To capture global context, ViM employs bidirectional scanning, processing sequences forward and backward to capture dependencies effectively \cite{zhu2024vision}. The model integrates information across the scan direction as if it had evolved continuously, although pixels are discrete samples, adjacent pixels are highly correlated through smooth shading, edges with consistent slopes, and textures with gradients. As ViM scans image patches sequentially, treated as continuous systems, they must be discretized to determine how the hidden latent state in state space models evolves between adjacent tokens.

\subsection{Baseline SSM Discretization Method}

\textbf{Zero-order hold (ZOH)} is predominantly used in converting continuous signals to discrete samples because it is easy to implement and ensures that the system behaves predictably by holding the sample value constant until the next time step. This leads to approximation errors with visual data because it assumes that the signal is constant between samples, with no slopes, no intermediate changes in textures, and no sudden jumps at patch boundaries, which may not accurately reflect the true nature of complex spatial data \cite{moir2022rudimentsignalprocessing} \cite{zhang2007comparisonzohfoh}.
This can result in such errors as edges getting smeared, textures losing sharpness, and spatial details drift across long sequences.

In this study, we replace the ZOH operation with more advanced discretization methods, each designed to better capture dynamic changes between samples while keeping the ViM architecture and state update equations structurally identical.
The general discrete-time state-space formulation remains:
\begin{equation}
    h_{t+1} = \bar{\mathbf{A}} h_t + \bar{\mathbf{B}} x_t, \quad y_t = \mathbf{C} h_t
\label{eq:state_update}
\end{equation}
with the ZOH approximating equation:
$$ y(t) = \sum_n x(nT) \cdot \text{rect}(\frac{t - nT - T/2}{T}) $$
where rect(t) is the rectangular function and the computation of $\bar{\mathbf{A}}$ and $\bar{\mathbf{B}}$ differs according to the chosen discretization method.

\subsection{Advanced Discretization Methods}

\textbf{First-order hold (FOH)} assumes that the input $x(t)$ varies linearly between two samples, rather than being held constant as in ZOH. This yields smoother transitions and better approximations for signals with continuous slopes\cite{moir2022rudimentsignalprocessing}\cite{zhang2007comparisonzohfoh}.
For a continuous system $\frac{dh(t)}{dt} = \mathbf{A}h(t) + \mathbf{B}x(t)$, under FOH, the discrete time equivalent is:

\begin{equation}
    \bar{\mathbf{A}} = e^{\mathbf{A}\Delta}
\end{equation}
\begin{equation}
    \bar{\mathbf{B}} = \mathbf{A}^{-2}(e^{\mathbf{A}\Delta} - \mathbf{I} - \mathbf{A}\Delta)\mathbf{B}
\end{equation}

This formulation preserves the first-order derivative of $x(t)$ within the sampling interval, resulting in a more accurate representation of input-driven dynamics, especially in systems with moderate rate-of-change in features.
\newline
\newline

\begin{figure}[t]
\centering
\includegraphics[width=0.7\linewidth]{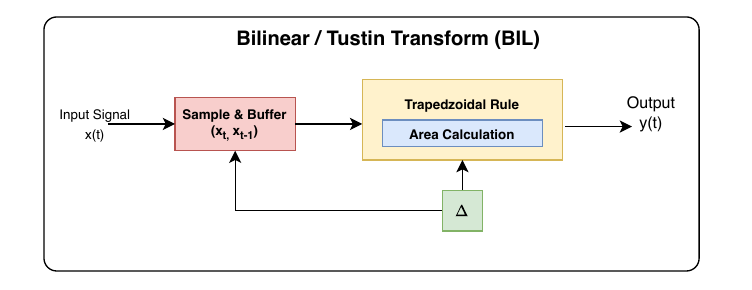}
\caption{BIL employs a trapezoidal integration rule to approximate the continuous input signal area, ensuring numerical stability and preserving high-frequency features during the discretization.}
\label{fig_bil}
\end{figure}

\textbf{The Bilinear/Tustin Transform (BIL)} provides a mapping between the continuous domain and the discrete domain, while minimizing frequency distortion. In vision tasks, the spatial frequency corresponds to the rate of change in pixel or feature values, where low frequencies capture smooth regions and global structure, and high frequencies represent edges, textures, and fine details. Preserving these high-frequency components during discretization is critical for accurate visual representation.
As illustrated in Fig.\ref{fig_bil}, BIL differs from hold-based methods by approximating the integration operation rather than the input signal itself. It employs a trapezoidal-rule formulation that uses both current and previous samples, treating the sampling period $\Delta$ as a fixed scaling factor for the calculation of the trapezoidal area rather than a dynamic interpolation variable as in FOH. 

When applied to a continuous system matrix, the discrete time representation is:
\begin{equation}
    \bar{\mathbf{A}} = (\mathbf{I} - \tfrac{\Delta}{2}\mathbf{A})^{-1}(\mathbf{I} + \tfrac{\Delta}{2}\mathbf{A})
\end{equation}
\begin{equation}
    \bar{\mathbf{B}} = (\mathbf{I} - \tfrac{\Delta}{2}\mathbf{A})^{-1}\Delta \mathbf{B} 
\end{equation}

Unlike FOH, which can "blur" high-frequency content through linear approximation, BIL preserves fine spectral variations while maintaining system stability and accurate frequency-domain behavior \cite{aastrom2013compcontrolsystems, chen2022improving, oppenheim2010discretetimesig}, making it particularly effective for ViM layers sensitive to fine spectral variations in texture and edges.
\newline


\begin{figure}[t]
\centering
\includegraphics[width=0.7\linewidth]{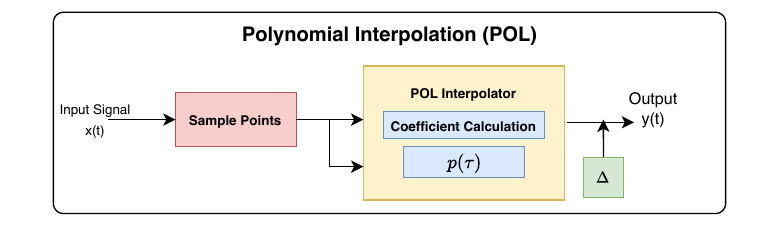}
\caption{POL fits a polynomial through sampled points of the input, allowing the discrete model to approximate smooth, non-linear transitions between samples.}
\label{fig_pol}
\end{figure}

\textbf{Polynomial interpolation (POL)} incorporates higher order terms as coefficients, improving the accuracy of the discrete approximation by capturing curvature in the input, resulting in smoother and more accurate latent state updates compared to lower order hold methods\cite{chen2025approxmethodsdiscretize}\cite{takacs2022hohmethods}\cite{ascher1998computermethodsode}. Fig.\ref{fig_pol} shows the block diagram that allows the discrete model to approximate smooth, non-linear transitions between samples.
The discrete equivalent of the state-space system is given by integrating the polynomial input over time:
\begin{equation}
    \bar{\mathbf{A}} = e^{\mathbf{A}\Delta}
\end{equation}
\begin{equation}
    \bar{\mathbf{B}} = \int_0^{\Delta} e^{\mathbf{A}\tau}\mathbf{B} p(\tau) d\tau
\end{equation}
where $p(\tau)$ is the chosen polynomial basis function (e.g., quadratic or cubic). For example, for cubic interpolation, a closed-form solution can be expressed as:
\begin{equation}
    \bar{\mathbf{B}} = \mathbf{A}^{-1}(e^{\mathbf{A}\Delta} - \mathbf{I})\mathbf{B} + \frac{1}{2}\mathbf{A}^{-2}(e^{\mathbf{A}\Delta} - \mathbf{I} - \mathbf{A}\Delta)\mathbf{B}
\end{equation}

POL uses both past data and future sample points, a non-causal approach aligning directly with the bidirectional scan mechanism used in ViM and other vision SSMs. This allows the system to "see" the entire sequence at once rather than guessing the trajectory based solely on past momentum. Other discretization methods only project the signal forward based on the previous sampled points and may overshoot; POL, however, ensures the discretization error is minimized across the entire visual context.
\begin{figure}[t]
\centering
\includegraphics[width=0.7\linewidth]{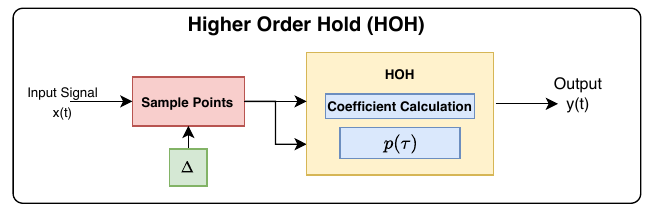}
\caption{HOH uses multiple samples to construct a higher-order polynomial approximation of the signal within the sampling interval, providing smoother reconstruction with potentially lower error than ZOH/FOH.}
\label{fig_hoh}
\end{figure}

\textbf{Higher-order hold (HOH)} generalizes both ZOH and FOH by modeling the input as a polynomial of $n^{\text{th}}$ order between samples. In contrast to POL, which applies fixed-order polynomial correction terms to improve local input smoothness. 
For an $n^{\text{th}}$ order hold, the discrete matrices are defined as:
\begin{equation}
    \bar{\mathbf{A}} = e^{\mathbf{A}\Delta}
\end{equation}
\begin{equation}
    \bar{\mathbf{B}} = \sum_{i=0}^{n} \mathbf{A}^{-(i+1)}\frac{(e^{\mathbf{A}\Delta} - \sum_{k=0}^{i} \frac{(\mathbf{A}\Delta)^k}{k!})}{i!} \mathbf{B}
\end{equation}
where,
\(n = 0\) corresponds to ZOH;
\(n = 1\) represents FOH;
\(n > 1\) denotes higher order (quadratic, cubic).
As shown in Fig.\ref{fig_hoh}, HOH offers superior accuracy for systems with rapidly changing inputs, such as changes in dynamic textures \cite{chen2025approxmethodsdiscretize}\cite{takacs2022hohmethods}.
\newline

\vspace{-6pt}
\begin{figure}[t]
\centering
\includegraphics[width=0.7\linewidth]{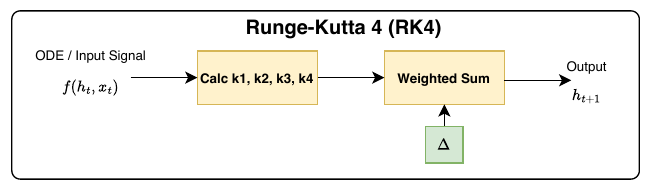}
\caption{RK4 is a fourth-order iterative method that approximates the value at the current time step determined by the previous time step value, where each increment is the product of the size of the interval and an estimated weighted average of the four slopes\cite{butcher1964rk4processes}\cite{ascher1998computermethodsode}.}
\label{fig_rk4}
\end{figure}

\textbf{Runge–Kutta 4 (RK4)} provides a fourth-order numerical integration scheme to approximate the continuous time dynamics of the system with high precision. For a given step size $\Delta$:
\begin{align}
    k_1 &= \mathbf{A}h_t + \mathbf{B}x_t \\
    k_2 &= \mathbf{A}(h_t + \tfrac{\Delta}{2}k_1) + \mathbf{B}x_{t+\Delta/2} \\
    k_3 &= \mathbf{A}(h_t + \tfrac{\Delta}{2}k_2) + \mathbf{B}x_{t+\Delta/2} \\
    k_4 &= \mathbf{A}(h_t + \Delta k_3) + \mathbf{B}x_{t+\Delta}
\end{align}

The next state is computed as:

\begin{equation}
    h_{t+1} = h_t + \tfrac{\Delta}{6}(k_1 + 2k_2 + 2k_3 + k_4)
\end{equation}


In ViM, RK4 discretization enables smoother propagation of state dynamics during training, as shown in Fig.\ref{fig_rk4}, especially for long-range dependencies or high-resolution images \cite{butcher1963rk4coefficients}\cite{butcher1964rk4processes}.
\subsection{Unified Hardware-Aware Discretization Framework}

The original Mamba addressed issues with SSM implementations being memory-bound due to frequent high-bandwidth memory (HBM) accesses by implementing a hardware-aware design that favors recomputation over storage to reduce bandwidth pressure. Discretization, state updates, and output projection are fused into a single GPU kernel, eliminating intermediate writes to HBM\cite{gu2023mamba}.

The original Mamba CUDA kernel only supported ZOH. We extend the hardware-aware algorithm to support all discretization methods evaluated in this work, ensuring identical fused execution, memory access patterns, and scan-based recurrence across methods.
Discretization is computed inside the selective scan kernel, keeping all intermediate values in registers and avoiding global memory writes. This design maintains optimized memory access and parallel scalability of the original Mamba implementation.

\section{Experiments}

This section outlines our experiments to integrate advanced discretization schemes into ViM. All experiments were trained according to the original ViM-Ti model in \cite{zhu2024vision} on 4 A100 GPUs. The continuous-to-discrete conversion is evaluated on three tasks: image classification, semantic segmentation, and object detection.
The resulting discrete-time parameters $\bar{\mathbf{A}}$ and $\bar{\mathbf{B}}$ replace the default ZOH matrices used in the state update function. 
The rest of the ViM pipeline from \cite{zhu2024vision} remains unchanged as shown in Fig.\ref{fig1}, ensuring comparability across all experiments.
\noindent
After discretization, the ViM layer updates the hidden state using Eq.~\eqref{eq:state_update}.

\subsection{Image Classification on ImageNet-1k \& Cifar100}

\textbf{Settings.} The original ViM model was trained on the ImageNet-1K dataset \cite{deng2009imagenetdb} and Cifar100 dataset \cite{krizhevsky2009cifar}. The top-1 and top-5 accuracy on the validation set are reported. Training settings were maintained according to \cite{touvron2021trainingdeit} with the same data augmentations defined in the original work (random cropping, random horizontal flipping, label-smoothing regularization, mixup). 

\textbf{Results.} 
Table~\ref{tab:image_classification} illustrates an improved, more efficient learning strategy for State Space Models to accurately classify different images. Using the same ViM model with no other architectural changes, BIL results in an accuracy of $71.95\%$, a $0.83\%$ improvement from the original ZOH method used in \cite{zhu2024vision}. This improvement, while not a large improvement, is achieved with approximately the same convergence of 232 epochs, indicating a more efficient numerical method integrated within the model.
The POL and HOH methods are the most accurate, while requiring more training epochs before reaching their maximum performance of $72.76\%$ and $72.74\%$, a $1.64\%$ and $1.62\%$, respectively.
To verify these results, the same comparison procedure was trained and evaluated on Cifar100 \cite{krizhevsky2009cifar}, showing a similar improved top-1 accuracy by approximately $2.32\%$ for the HOH compared to ZOH, and $0.85\%$ improvement for BIL.
The results demonstrate that advanced discretization methods can be applied to existing models without changing any other model parameters or architecture to reduce approximation errors, and extract stronger visual representations.
It should be noted that the original ZOH results were recreated and compared against non-fine-tuned models.

\begin{table*}[t]
    \centering
    \small
    \caption{Discretization methods results on image classification tasks.}
    \resizebox{\textwidth}{!}{
    \begin{tabular}{|c|c|c|c|c|c|c|c|}
        \hline
        \textbf{Discretization Method} & \textbf{Dataset} & \textbf{Training Time (dd-hh:mm)} & \textbf{Best Epoch / 300} & \textbf{Top-1 ACC} & \textbf{Top 5 ACC} \\
        \hline
         ZOH &  & 22-11:36 & 227 & 71.12 & 86.71 \\
         FOH &  & 22-11:36 & 226 & 71.14 & 86.49 \\
         BIL & ImageNet-1k & 22-18:27 & 232 & 71.95 & 90.75 \\
         POL &  & 23-16:31 & 272 & 72.76 & 91.21 \\
         HOH &  & 23-21:51 & 289 & 72.74 & 91.20 \\
         RK4 &  & 28-06:22 & 298 & 71.63 & 90.10 \\
        \hline
         ZOH &  & 0-01:29 & 225 & 43.84 & 71.23 \\
         FOH &  & 0-01:08 & 227 & 43.08 & 70.43 \\
         BIL & CIFAR100 & 0-01:28 & 230 & 44.69 & 71.44 \\
         POL &  & 0-01:38 & 271 & 45.87 & 72.23 \\
         HOH &  & 0-01:38 & 281 & 46.16 & 72.53 \\
         RK4 &  & 0-01:47 & 290 & 44.27 & 71.09 \\
        \hline
    \end{tabular}
    }
    \label{tab:image_classification}
\end{table*}

\subsection{Semantic Segmentation on ADE20K}

\begin{table*}[t]
    \centering
    \small
    \caption{Discretization methods results on semantic segmentation using ADE20K and object detection using MSCOCO.}
    \small
    \begin{tabular}{|c|c|c|c|c|c|c|c|}
        \hline
        \textbf{Discretization Method} & \textbf{\;\;mIoU\;\;} & \textbf{\;\;AP$^{box}_{75}$}\;\;\\
        \hline
        ZOH & 41.0 & 49.6 \\
        FOH & 40.2 & 49.4\\
        BIL & 43.9 & 52.1\\
        POL & 45.4 & 53.2 \\
        HOH & 45.2 & 54.1\\
        RK4 & 44.1 & 52.0\\
        \hline
    \end{tabular}
    \label{tab:downstream}
\end{table*}

\textbf{Settings.} 
We evaluate the proposed discretization methods on the ADE20K dataset \cite{zhou2019semanticade20k}, which contains 150 fine-grained semantic categories with 20K training images, 2K validation images, and 3K test images. All experiments are conducted using ViM as the backbone, integrated into the UperNet framework \cite{xiao2018upernet}.

The models were optimized using AdamW with a weight decay of $0.01$ and a total batch size of $16$. The learning rate was initialized at $6 \times 10^{-5}$ and decayed linearly over a total of $160$K iterations, with a linear warmup of $1{,}500$ iterations. Data augmentation includes random horizontal flipping, random rescaling, and random distortion.
Across all experiments, the ViM architecture, training schedule, and segmentation head remain unchanged. Only the discretization method used to compute the discrete-time SSM parameters is modified, ensuring that observed differences arise solely from the numerical discretization strategy.

\textbf{Results.}
Table~\ref{tab:downstream} reports the mean Intersection over Union (mIoU) results on ADE20K. BIL improves segmentation performance to $43.9$ mIoU, representing a $2.9$ point gain over the ZOH baseline used in previous ViM implementations \cite{zhu2024vision}. This improvement is achieved without changes to the model capacity or the training procedure, indicating that advanced discretization methods directly benefit segmentation tasks.
The POL and HOH methods achieve the strongest segmentation performance, with mIoU scores of $45.4$ and $45.2$, corresponding to gains of $4.4$ and $4.2$ over ZOH, respectively. These results suggest that higher-order approximations better capture fine-grained spatial transitions and boundary information for semantic segmentation. RK4 also improves over ZOH, although minimally, while FOH underperforms the baseline. 

\subsection{Object Detection on COCO2017}

\textbf{Settings.} 
We evaluated object detection performance on the MS COCO 2017 benchmark \cite{lin2014microsoftcocodataset}, which consists of 118K training images, 5K validation images, and 20K test images. All experiments use ViM as the backbone within the Cascade Mask R-CNN framework \cite{cai2019cascadercnn}. Unlike ViT-based backbones, ViM is used directly without architectural modifications for high-resolution processing.

Training is performed using the AdamW optimizer with a weight decay of $0.1$ and a total batch size of $64$. The learning rate is initialized at $1 \times 10^{-4}$ and decayed linearly over $380$K iterations. Large-scale jitter enhancement is applied, resizing the input images to $1024 \times 1024$ during training. During evaluation, the images are resized so that the shorter side is $1024$ pixels. All detection and training hyperparameters are kept constant across discretization methods.

\textbf{Results.}
Detection results are summarized in Table~\ref{tab:downstream} using AP$^{box}{75}$ as the evaluation metric. BIL improves AP$^{box}{75}$ from $49.6$ under ZOH to $52.1$, yielding a gain of $2.5$. POL and HOH methods achieve AP$^{box}_{75}$ scores of $53.2$ and $54.1$, respectively. These gains indicate that higher-order discretization more effectively preserves spatial dynamics relevant for object localization and bounding box refinement.
Consistent with the classification and segmentation results, FOH does not offer a significant improvement over ZOH, while RK4 provides moderate gains but does not match the performance of POL or HOH. The results confirm that the discretization strategy has a substantial impact on downstream detection accuracy.

\subsection{Statistical Significance}

\begin{figure}[t]
\centering
\includegraphics[width=0.8\linewidth]{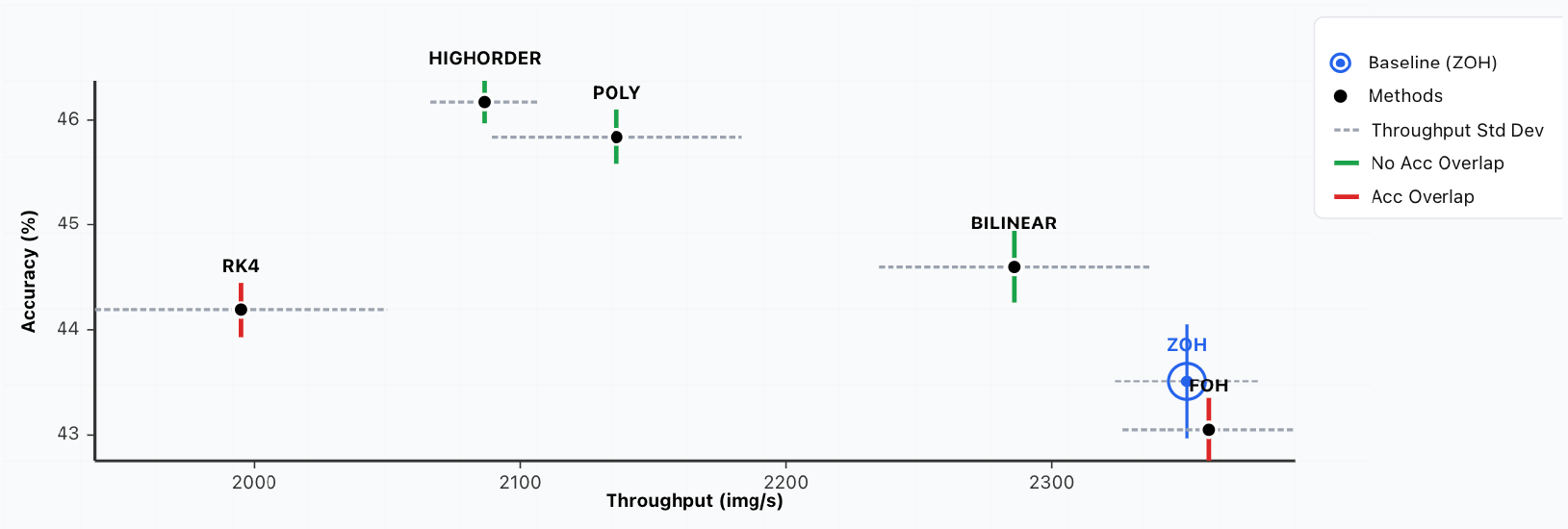}
\caption{Statistical scatter plot - performance vs. efficiency. Mean ± standard deviation (std) of different methods.}
\label{fig_stats}
\end{figure}

Further experiments shown in Fig.\ref{fig_stats} evaluate the six discretization methods across 10 random seeds to determine whether the accuracy improvements were statistically significant. A one-sided paired permutation test was used to assess whether each method improved accuracy relative to the baseline ZOH \cite{zhang2007comparisonzohfoh}. A significance level of $\alpha = 0.05\%$ was selected and the accuracy gains were considered significant only if they exceeded $0.70\%$, which corresponds to the upper limit of the typical seed-induced variation on CIFAR-100 \cite{krizhevsky2009cifar}. Throughput was observed to have minor variations relative to the complexity of each method, and accuracy gain did not degrade computational efficiency.

BIL demonstrated a statistically significant improvement in accuracy of $1.09\%$ ($\geq0.70\%$ threshold, p<0.001) with an observed downtime of only $2.77\%$.
HOH showed a statistically significant improvement in accuracy of $2.66\%$ (p<0.001), well above the threshold of $0.70\%$, with an observed decrease in performance of $11.23\%$. This decrease is expected as it follows the increasing order of computational operations required to solve each advanced discretization method.
POL achieved a statistically significant improvement in accuracy of $2.33\%$ (p<0.001) with an observed decrease in throughput of $9.13\%$.
FOH and RK4 did not meet the required accuracy threshold and therefore do not offer statistically significant improvements. FOH improved accuracy by only $0.48\%$ (below the threshold of $0.70\%$) and did not show statistical significance (p=0.97). RK4 improved accuracy by only $0.68\%$ (p<0.001), also below the practical significance threshold.
BIL, HOH, and POL provide statistically significant accuracy gains, with modest throughput trade-offs.


\section{Discussion}
\subsection{Peak Performance and Training Convergence}

Beyond the final precision, we observe substantial differences in the convergence speed across discretization methods, represented as the training time in Table~\ref{tab:image_classification}. These differences arise from both the order of the numerical approximation and the causality structure of each method.

POL exhibits slower convergence despite using a relatively low approximation order because it is non-causal. The bidirectional formulation introduces dependencies on both past and future states, causing gradients to propagate in multiple directions. This requires additional training iterations for the model to reconcile competing gradient signals and achieve stable performance.
HOH, while fully causal, relies on higher-order approximations, introducing a stronger coupling between model parameters. As a result, gradients become more sensitive to small parameter changes, and optimization proceeds more cautiously. This results in slower convergence, even though these methods provide greater final accuracy.
RK4 presents the most challenging optimization dynamics. Its multi-stage weighted average algorithm produces deep gradient paths and strong interactions among intermediate terms. This increases the risk of gradient instability and limits practical gains in accuracy under standard training regimes, consistent with its failure to surpass the practical significance threshold.

These observations highlight a fundamental trade-off between convergence speed and accuracy. Lower-order, causal discretizations favor faster optimization; however, higher-order and non-causal methods improve final performance at the cost of increased training complexity. From a practical standpoint, the bilinear transform emerges as a strong default choice, offering statistically significant gains with minimal computational overhead and fast convergence. 


\subsection{Inference Speed}

\begin{table*}[t]
    \centering
    \small
    \caption{Fastest method per batch size based on latency and throughput.}
    {
    \begin{tabular}{|l|c|c|c|}
        \hline
        Fastest Method & Batch Size & Latency (ms/img) & Throughput (img/s) \\
        \hline
        BIL & 1 & 13.46 & 74.28 \\
        POL & 4 & 3.39 & 298.16 \\
        FOH & 8 & 1.70 & 587.08 \\
        FOH & 16 & 0.86 & 1164.25 \\
        POL & 32 & 0.48 & 2193.21 \\
        ZOH & 64 & 0.42 & 2257.83 \\
        POL & 128 & 0.38 & 2588.37 \\
        FOH & 256 & 0.32 & 3272.25 \\
        BIL & 512 & 0.21 & 4804.37 \\
        \hline
    \end{tabular}
    }
    \label{tab:fastest}
\end{table*}

\begin{table*}[t]
    \centering
    \small
    \caption{Summary of deployment scenarios.}
    \resizebox{\textwidth}{!}{
    \begin{tabular}{|l|c|c|c|c|c|}
        \hline
        Deployment Scenario & Batch Size & Best Method & Performance & Improvement Over Baseline (ZOH) \\
        \hline
        Real-Time/Edge & 1 & BIL & 13.46 ms (Latency) & ~0.37\% Faster \\
        High-Throughput / Cloud & 512 & BIL & 4,804 img/s (Throughput) &  ~0.67\% Faster \\
        Balanced / Batched & 32 & POL & 2,193 img/s (Throughput) & \textit{Context Dependent} \\
        \hline
    \end{tabular}
    }
    \label{tab:summary}
\end{table*}

Tables~\ref{tab:fastest} and~\ref{tab:summary} show our benchmarked inference speeds across batch sizes to evaluate the efficiency of each advanced discretization method. BIL establishes an improvement in efficiency, outperforming the other methods in both low-latency (with batch size 1) and high-throughput comparisons (with batch size 512). As shown in Table~\ref{tab:summary}, BIL achieves the lowest latency of 13.46 ms/img at batch size 1, making it the optimal choice for real-time applications. At batch size 512, it maximizes GPU saturation with a peak throughput of 4,804 img/s. While POL and FOH offer marginal gains at intermediate batch sizes (4–256), BIL provides the most robust solution for scalable deployment.

For latency-critical applications (batch size 1), BIL is the superior discretization strategy at 13.46 ms, providing a distinct advantage over higher-order methods which may incur computational overhead without throughput gains at low saturation.
As batch sizes increase to 512, BIL shows advantageous throughput (4,804 img/s), suggesting that it handles memory-bound and compute-bound transitions more effectively than the other methods.
Intermediate batch sizes (4 to 256) reveal a volatile landscape where POL and FOH frequently outperform BIL and the other methods, as shown in Table~\ref{tab:fastest}. For example, at batch size 32, POL peaks at 2,193 img/s. This suggests that our specific hardware kernel optimizations favor polynomial approximations at medium loads.

The advanced discretization methods defined in this work increase numerical accuracy by introducing additional matrix operations; however, they incur almost equal or slightly better inference cost than ZOH, with practical latency differences typically small. This indicates a minimal tradeoff for deploying a more accurate model in scalable environments.

\section{Conclusion}\label{sec:chall}

This study demonstrates that incorporating advanced discretization strategies yields measurable performance gains in visual tasks without any other changes to model parameters or architectural design. 
POL and HOH methods provide the highest accuracy, and BIL offers the most practical balance between precision and computational efficiency in classification, segmentation, and detection. BIL ensures that spatial frequencies encoded in image patches evolve correctly in the latent state, preserving edges, textures, and long-range accuracies.
Our comprehensive comparison of discretization strategies in SSM for vision establishes a clear direction for future SSM research by using more advanced methods 
\subsubsection{Acknowledgements}
This research was partly funded by the Natural Sciences and Engineering Research Council of Canada (NSERC) and the  Canada Foundation for Innovation (CFI).


\balance

\end{document}